\pdfoutput=1

\documentclass[letterpaper, 10 pt, conference]{ieeeconf}  %

\IEEEoverridecommandlockouts                              %

\usepackage[T1]{fontenc}                                                          

\usepackage{tikz} %

\usepackage{amsfonts}	%
\usepackage{amsmath}	%
\usepackage{amssymb}    %
\usepackage{siunitx}
\usepackage{pifont}   %
\usepackage{xspace}  

\usepackage{svg}

\usepackage{booktabs}
\usepackage{makecell}  %
\usepackage[flushleft]{threeparttable}  %
\usepackage{multirow}
\usepackage{rotating}

\usepackage{xspace}    %
\usepackage{xcolor}    %
\let\labelindent\relax
\usepackage[inline]{enumitem} %
\usepackage{textcomp}  %
\usepackage{gensymb}   %
\usepackage{graphicx} %
\usepackage{microtype}
\usepackage{cite}
\usepackage{flushend}
\usepackage{cuted}

\usepackage{pgfplots}
\pgfplotsset{width=7cm, compat=1.10}
\usepgfplotslibrary{fillbetween}
\usepackage{tabularx}

\newcolumntype{P}[1]{>{\centering\arraybackslash}p{#1}}

\makeatletter
\let\NAT@parse\undefined
\makeatother

\usepackage{url}
\usepackage[pdfencoding=auto, breaklinks,colorlinks,hypertexnames=false]{hyperref} %
\usepackage[hang,flushmargin]{footmisc}

\usepackage[capitalize]{cleveref}
\crefname{section}{Sec.}{Secs.}
\Crefname{section}{Section}{Sections}
\Crefname{table}{Table}{Tables}
\crefname{table}{Tab.}{Tabs.}

\usepackage[nogroupskip,acronyms,nopostdot,style=super,nonumberlist,toc]{glossaries}

\usepackage{colortbl}

\newcommand{\ours}{QueryArt\xspace}

\usepackage{amsmath}
\usepackage{amssymb}

\DeclareMathAlphabet{\mathcal}{OMS}{cmsy}{m}{n}

\usepackage{xspace}

\renewcommand{\[}{\begin{equation}}
\renewcommand{\]}{\end{equation}}

\renewcommand{\baselinestretch}{0.99}

\definecolor{red}{RGB}{255, 0, 0}   %
\definecolor{orange}{RGB}{255, 77, 0}   %
\definecolor{green}{RGB}{0, 128, 0}   %
\definecolor{purple}{RGB}{160, 32, 240}   %
\definecolor{lightblue}{RGB}{52, 155, 235}   %
\definecolor{darkmagenta}{RGB}{204, 51, 139} %

\usepackage[capitalize]{cleveref}

\crefname{figure}{Fig.}{Figs.}
\Crefname{figure}{Figure}{Figures}
\crefname{section}{Sec.}{Secs.}
\Crefname{section}{Section}{Sections}
\Crefname{table}{Table}{Tables}
\crefname{table}{Tab.}{Tabs.}
\crefname{algorithm}{Algo.}{Algos.}
\Crefname{algorithm}{Algorithm}{Algorithms}
\crefname{appendix}{Sec.}{Secs.}
\Crefname{appendix}{Section}{Sections}

\usepackage{graphics} %
\usepackage{svg}

\title{\LARGE \bf
Query-Conditioned Articulation Estimation from a Single Image
}

\newif\ifanonymous

\ifanonymous
\author{Anonymous submission}
\else
\author{Abdelrhman Werby, Fabio Scaparro, Kai O. Arras%
\thanks{A. Werby, F. Scaparro and K. O. Arras are with the Socially Intelligent Robotics Lab, Institute for Artificial Intelligence, University of Stuttgart, Germany. Email: {\tt\small abdelrhman.werby@ki.uni-stuttgart.de}}%
\thanks{The authors thank the International Max Planck Research School for Intelligent Systems (IMPRS-IS) for supporting Abdelrhman Werby.}
}
\fi

\begin{document}

\bstctlcite{IEEEtran:BSTcontrol}

\maketitle
\thispagestyle{empty}
\pagestyle{empty}

\suppressfloats[t]

\begin{abstract}
    Enabling robots to estimate the kinematic parameters of articulated objects unlocks a wide range of capabilities for interaction and manipulation. The estimation has to happen from the information the robot currently observes, often just a single RGB image of an object it has never seen before.
Current single-image approaches couple articulation part segmentation with articulation estimation, making their predictions vulnerable to missed detections and incorrect part associations, and they regress metric 3D geometry that a single view fixes only up to scale.
We present \ours, a model that estimates articulation parameters from a single RGB image, a 2D query point, and camera intrinsics. \ours is trained to estimate the 3D articulation geometry relative to the queried point and in units of its depth, which keeps its target identifiable from the image alone. A single depth measurement at the query point then supplies the scale and recovers the metric parameters. We train \ours on a curated mixture of synthetic and real-world articulation datasets. We evaluate \ours on several benchmarks and compare it against recent baselines. \ours outperforms recent baselines on most articulation metrics, including on out-of-distribution data. To demonstrate the model's capabilities in real-world settings, we evaluate \ours on a mobile manipulator across 57 manipulation trials spanning 16 object parts and five viewpoint classes, achieving a 70.2\% success rate. We provide code and videos at: \url{https://abwerby.github.io/queryart/}.

\end{abstract}
\glsresetall

\begin{figure}[t]
    \centering
    \includegraphics{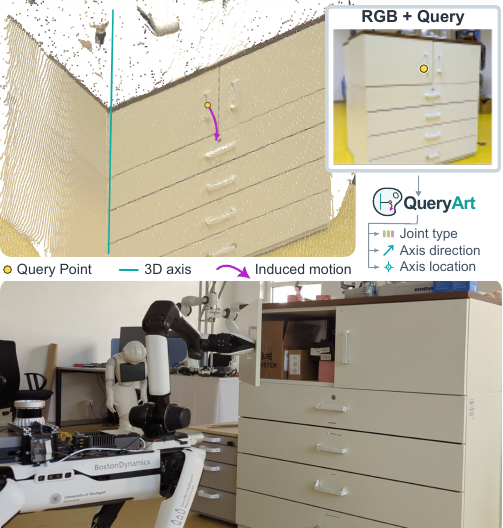}
    \caption{\ours estimates articulation in a normalized coordinate space from a single RGB image and a 2D query point. Depth at the query point enables lifting the prediction to metric 3D space. The magenta arrow illustrates motion consistent with the predicted articulation. Bottom: real-world deployment on a mobile manipulator.}
    \label{fig:teaser}
\end{figure}

\section{Introduction}
\label{sec:introduction}

Robots deployed in human-centered environments must interact with articulated objects such as doors, cabinets, drawers, and appliances~\cite{zhang2023flowbotpp, gupta2025opening}. Successfully manipulating such objects requires estimating their underlying kinematic model, including the joint type, motion axis, and axis location. These parameters support a wide range of downstream capabilities, from navigating through closed doors to searching for objects inside cabinets and drawers. Ideally, a robot should infer them for an unseen object before physical interaction, using only the observations available from its current viewpoint.

Single-view articulation estimation methods~\cite{jiang2022opd, sun2024opdmulti, li2024mopd, pmlr-v270-huang25b} jointly detect and segment movable parts and estimate their motion parameters from a single image. This formulation couples articulation estimation with part localization; as a result, a localization failure propagates directly to the kinematic prediction. Recent approaches~\cite{Chen_2025_CVPR, zhang2026molmomotion} instead predict a 3D interaction trajectory from a single image using a learned motion prior, but their output is not parameterized by an explicit revolute or prismatic joint model and therefore does not guarantee kinematic consistency by construction. Both families of methods regress metric 3D geometry directly from the image, although a single view determines the scene only up to scale.
Methods such as USDNet~\cite{Halacheva_2025_ICCV} operate on complete 3D scene point clouds and jointly predict articulated-part segmentation and motion attributes; however, they require prior scene exploration. Video-based methods~\cite{pmlr-v305-werby25a, buechner2026momasg} infer articulation by observing the articulated object in motion, requiring the object to be manipulated or observed in multiple states before its articulation can be estimated.

To this end, we present \ours, a model for estimating articulation parameters from a single calibrated RGB image and a 2D query point on the articulated part (\cref{fig:teaser}). It predicts the joint type, the 3D motion axis, and the query-relative axis location, normalized by the query point's depth. This normalization makes the target invariant to uniform scaling of the scene. At inference time, obtaining the query point's depth provides the metric scale, removing the need for the model to infer absolute size from appearance. Conditioning on an interaction point decouples the question of \emph{where to interact} from the question of \emph{how the selected point moves}, allowing the query to originate from a human click or an independent affordance-grounding system~\cite{pmlr-v270-yuan25c, deitke2025molmo}.
The model combines a frozen visual foundation encoder~\cite{simeoni2025dinov3}, deformable attention~\cite{zhu2021deformabledetr}, and lightweight articulation-specific heads. \ours is trained on a curated mixture of two synthetic and three real-world articulation datasets. When the query point metric depth is available, a deterministic lifting step recovers the metric 3D motion axis.
Across held-out splits of four training datasets and two datasets excluded from training entirely, \ours improves over recent single-image baselines on most articulation metrics, and 57 trials on a mobile manipulator show that its predictions are accurate enough to open furniture the model has never seen, at a 70.2\% success rate.   

Our main contributions are:
\begin{itemize}
    \item A query-conditioned formulation that models articulation estimation as regression of normalized articulation geometry, given a single calibrated RGB image and a 2D query point.
    \item An efficient architecture that combines high-resolution frozen visual foundation features and lightweight prediction heads with a compact, physically structured axis-hinge representation.
    \item An extensive evaluation on synthetic and real-world articulation benchmarks, including out-of-distribution data, demonstrating improved articulation estimation over recent baselines.
    \item Real-robot experiments showing that the predicted parameters support analytic motion generation and manipulation of previously unseen objects.
\end{itemize}

\section{Related Work}
\label{sec:related}

\subsection{Single RGB-based Articulation Estimation}

Single-image articulation estimation, such as OPD and OPDMulti~\cite{jiang2022opd, sun2024opdmulti}, jointly detects openable parts and regresses their joint type, motion axis, and origin, coupling articulation estimation with part localization. 3DOI~\cite{Qian_2023_ICCV} predicts interaction properties at a queried image location, including a projected 2D rotation axis, but does not recover a complete 3D articulation model. CARTO~\cite{heppert2023carto} reconstructs the shape, pose, and joint state of multiple articulated objects from a single stereo RGB observation, but represents each object as a whole rather than conditioning on a queried interaction point. A3VLM~\cite{pmlr-v270-huang25b} uses vision-language models to predict joint type and 3D motion axes from object-part boxes or textual descriptions, while other methods~\cite {Chen_2025_CVPR, zhang2026molmomotion} directly predict 3D interaction trajectories from a single image using learned motion priors. Such methods do not explicitly parameterize motion as revolute or prismatic and therefore do not enforce the corresponding kinematic structure by construction.

\subsection{Articulation Estimation from 3D Observations}

Several approaches estimate articulation directly from 3D geometry. ANCSH~\cite{li2020ancsh} estimates category-level part poses and joint parameters from a single depth image using a normalized coordinate hierarchy, while ScrewNet~\cite{jain2021screwnet} predicts category-independent articulation models from depth observations using a screw-theoretic parameterization, and Zeng et al.~\cite{zeng2021visual} identify the articulation mechanism between part pairs from a single RGB-D frame. FlowBot++~\cite{zhang2023flowbotpp} is the closest to our geometric parameterization, as it regresses the perpendicular displacement from every point of an object to the joint axis, which is the same quantity our model predicts at the query point. However, it operates on a point cloud and relies on a part segmentation mask, and it regresses the displacement in metric scale, which a single image cannot determine. DUST-Net~\cite{jain2022dustnet} predicts distributions over articulation parameters from depth observations, explicitly modeling uncertainty over joint type, axis, and configuration. More recent scene-level methods~\cite{Halacheva_2025_ICCV} operate on reconstructed 3D point clouds to jointly segment articulated parts and estimate their motion parameters. Other approaches jointly recover object geometry and articulation from observations of multiple object states. Ditto~\cite{Jiang_2022_CVPR} reconstructs articulated objects and estimates their kinematic models from observations before and after interaction, and Weng et al.~\cite{weng2024digitaltwin} build digital twins of unknown articulated objects from two RGB-D scans. PARIS~\cite{liu2023paris}, ArtGS~\cite{liu2025artgs}, and ScrewSplat~\cite{kim2025screwsplat} respectively optimize a part-level radiance, Gaussian, or screw-based representation jointly with the articulation parameters from multi-view observations of different joint states. While these methods benefit from explicit 3D geometry, they require depth, reconstructed scenes, or observations of multiple configurations, limiting their applicability when a robot encounters an unseen object from a single view.

\subsection{Multiple-Observation Articulation Estimation}

A complementary line of work infers articulation by observing how an object moves. Sturm et al.~\cite{sturm2011probabilistic} fit and select rigid, revolute, and prismatic kinematic models from sequences of observed relative part poses, performing system identification after motion has been observed and thereby resolving ambiguities that appearance alone leaves open. Interactive-perception systems obtain this motion by manipulating the object themselves~\cite{katz2013interactive, pillai2014learning, martinmartin2016integrated}, which Ditto in the House~\cite{hsu2023dittohouse} scales to room-scale scenes, while factor-graph formulations~\cite{heppert2022catind, buchanan2024online} track articulated parts online and fuse visual with proprioceptive measurements. ArtiPoint~\cite{pmlr-v305-werby25a} instead tracks parts through egocentric RGB-D sequences of human interaction. Although these methods provide accurate kinematic estimates from observed motion, they require multiple observations or an interaction trajectory before the articulation can be inferred. In contrast, we predict plausible articulation hypotheses before interaction from a single monocular observation.

\section{Approach}
\label{sec:approach}

\begin{figure*}[t]
    \centering
    \includegraphics[width=\textwidth]{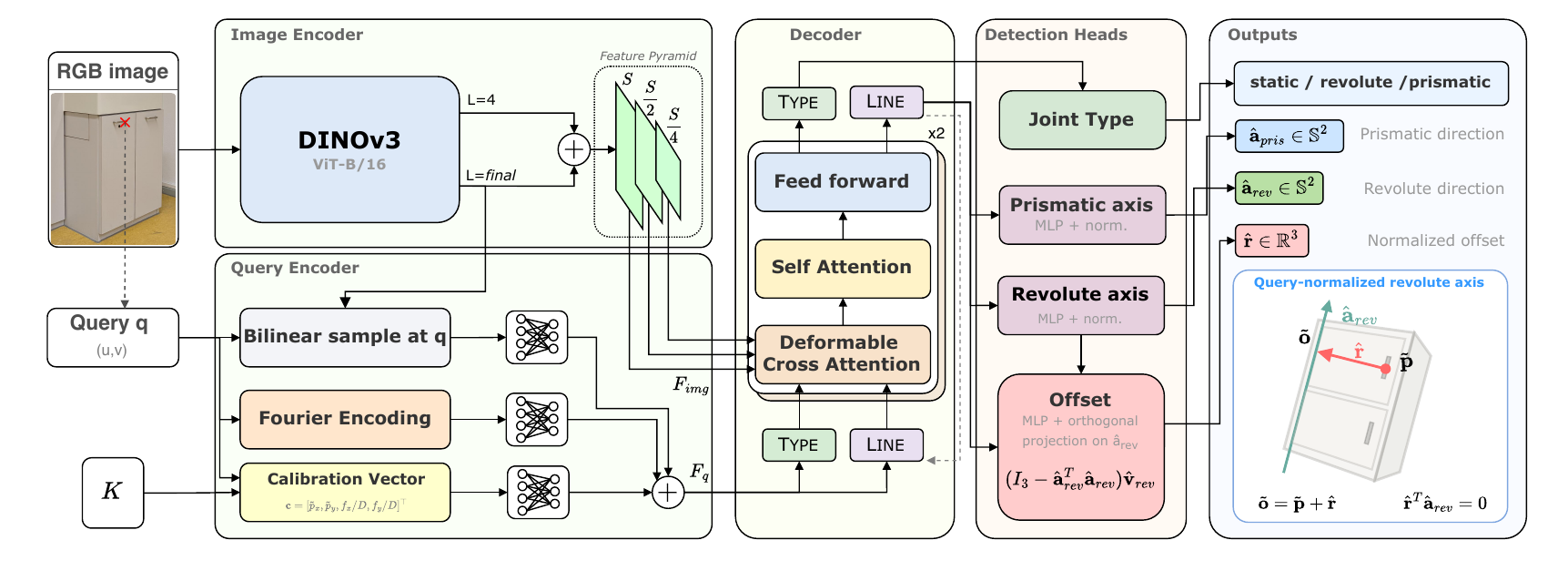}
    \caption{QueryArt architecture. A frozen vision transformer (DINOv3 ViT-B/16) encodes the RGB image; early and final-block features are fused into a three-scale pyramid $F_{\mathrm{img}}$. The query embedding $F_q$ is built from the same point-wise visual features sampled at the query, a Fourier encoding of the pixel location, and a hand-crafted calibration vector derived from the intrinsics $K$. $F_q$ is added to two randomly initialized learned tokens: \textsc{Type} and \textsc{Line}, which then pass through two transformer blocks that cross-attend to $F_{\mathrm{img}}$ and self-attend to each other. The prediction heads output the joint classification and the axis parameters for both moving types. Right: the predicted geometry in query-normalized coordinates. The offset $\hat{\mathbf{r}}$ is orthogonal to $\hat{\mathbf{a}}_{\mathrm{rev}}$, so it points from the unprojected query $\tilde{\mathbf{p}}$ to the nearest point $\tilde{\mathbf{o}} = \tilde{\mathbf{p}} + \hat{\mathbf{r}}$ on the revolute axis. Expressing $\hat{\mathbf{r}}$ in units of the query depth makes the axis recoverable up to a single global scale factor.}
    \label{fig:method_overview}
\end{figure*}

\subsection{Problem Formulation}
\label{sec:approach:formulation}

Given a single RGB image $I$, a pixel coordinate query $\mathbf{q}=(u,v)$, and camera intrinsics matrix $K$, our goal is to estimate the articulation parameters of the queried point. We predict a joint type $j\in\{\text{static},\text{revolute},\text{prismatic}\}$. For static parts, no further geometry is needed. For moving parts a unit direction $\hat{\mathbf{a}}\in\mathbb{S}^{2}$ describes the axis of motion. The direction is identified up to sign: $\mathbf{a}$ and $-\mathbf{a}$ describe the same unoriented axis. The direction fully specifies a prismatic articulation, therefore we just predict $\hat{\mathbf{a}}_{\mathrm{pris}}$. A revolute part additionally needs the origin of the joint axis. In particular, the axis is a 3D line
\begin{equation}
    \ell(\mathbf{a},\mathbf{o})
    =\{\mathbf{o}+\lambda\mathbf{a}\mid\lambda\in\mathbb{R}\},
    \label{eq:axis_line}
\end{equation}
where the joint origin $\mathbf{o}$ is any point on the line.

Let $\mathbf{p}=[x_q, y_q, z_q]$ be the unprojected query point and let $\mathbf{o}_{\perp}$ be the projection of $\mathbf{p}$ on $\ell(\mathbf{a},\mathbf{o})$. Among the infinitely many joint origins, $\mathbf{o}_{\perp}$ is the closest to $\mathbf{p}$. Together with the axis $\hat{\mathbf a}_{\mathrm{rev}}$, the network predicts the offset $\mathbf r$ such that:
\begin{equation}
    \mathbf r =\mathbf o_\perp - \mathbf{p},
    \qquad
    \mathbf r^{\top}\hat{\mathbf a}_{\mathrm{rev}} = 0.
    \label{eq:offset_target}
\end{equation}
This allows the network not to commit to a specific joint origin.

Since a single RGB image cannot identify $\ell(\mathbf{a},\mathbf{o})$ on a metric scale, we remove the dependency on depth information by expressing the axis offset $\mathbf r$ in a space normalized by the query depth. Let $\tilde{\mathbf{p}}=K^{-1}[u,v,1]^{\top}$ be the unprojected query point at unit depth, so that $\mathbf{p}=z_q\tilde{\mathbf{p}}$.
Then the predicted offset in normalized coordinates is $\hat{\mathbf r} = {\mathbf r }/{z_q} $.
The predicted revolute axis in normalized camera coordinates is
\begin{equation}
    \tilde{\ell}
    = \left\{
    \tilde{\mathbf{o}}
    +\lambda\hat{\mathbf{a}}_{\mathrm{rev}}
    \mid\lambda\in\mathbb{R}
    \right\},
    \label{eq:normalized_axis_line}
\end{equation}
where:
\begin{equation}
    \tilde{\mathbf{o}}
    =\tilde{\mathbf{p}}+\hat{\mathbf{r}},
    \qquad
    \hat{\mathbf{r}}^{\top}\hat{\mathbf{a}}_{\mathrm{rev}}=0.
    \label{eq:hinge_offset}
\end{equation}
Here, $\tilde{\mathbf{o}}=\mathbf o_\perp / z_q$ is normalized by the query depth, and $\hat{\mathbf{r}}$ points from the normalized query position to the nearest point on the predicted axis line $\tilde{\ell}$ and perpendicular to the axis direction $\hat{\mathbf{a}}_{\mathrm{rev}}$. Although expressed using three coordinates, $\hat{\mathbf{r}}$ has only two degrees of freedom once $\hat{\mathbf{a}}_{\mathrm{rev}}$ is fixed. This representation specifies a complete 3D revolute axis up to global scale.
During training, the ground-truth query depth is used to construct the normalized geometric targets in~\cref{eq:normalized_offset_target}. During inference, if metric query depth $z_q$ is available, we can recover the metric-scale 3D axis line position as $\hat{\mathbf{o}}=z_q\tilde{\mathbf{o}}$, while its direction $\hat{\mathbf{a}}_{\mathrm{rev}}$ remains unchanged.
 
\subsection{Model Architecture}
\label{sec:approach:architecture}
\ours's architecture is illustrated in \cref{fig:method_overview}.
Given a single RGB image $I$, a pixel query $\mathbf{q}=(u,v)$, and camera intrinsics $K$, we proceed as follows:

We start by extracting image features using the frozen DINOv3 ViT-B/16 image encoder~\cite{simeoni2025dinov3}, producing 768-channel patch features with a stride of 16. Following~\cite{ranftl2021dpt}, we keep early features in block 4 and late features in the final block of the transformer model. Then we add the two features, weighted by a learned sigmoid scalar, and project them to 256 channels.
Estimating articulation parameters requires the model to locate fine-grained details and understand the global semantic context; therefore, we construct a three-scale feature pyramid~\cite{lin2017fpn} using three sequential $3\times3$ convolutional blocks, downsampling the fused features $F_{\mathrm{img}} \in \mathbb{R}^{S \times S \times 256}$ to resolutions $S$, $S/2$, and $S/4$, respectively.

We create a query embedding that combines three complementary signals. First, we normalize the pixel query to $\bar{\mathbf{q}}\in[0,1]^2$, then encode it using a Fourier positional encoding~\cite{mildenhall2020nerf}. Second, we bilinearly sample the final DINOv3 block feature at the query location, giving a 768-channel descriptor. Third, we construct the calibration vector
\begin{equation}
    \mathbf{c}=\left[\tilde p_x,\ \tilde p_y,\ \frac{f_x}{D},\ \frac{f_y}{D}\right]^{\top},
    \qquad D=\max(W,H),
    \label{eq:camera_embedding}
\end{equation}
where $W$ and $H$ are the width and height of the source image. Each of the three signals is projected into a 256-dimensional embedding by an MLP, then summed to produce $F_{q} \in \mathbb{R}^{256}$.
The query embedding $F_q$ is added to two randomly initialized learned task tokens, producing a \textsc{Type} token and a \textsc{Line} token. Two decoder layers update them: first, the tokens are updated through deformable cross-attention~\cite{zhu2021deformabledetr} to the three-scale image pyramid; then, self-attention is applied between the Type and Line tokens, followed by a token-wise feed-forward network.

\subsection{Articulation Heads}
\label{sec:approach:heads}

We use four output heads. A single MLP classifier reads the \textsc{Type} token and predicts the joint type $p(j\mid I,\mathbf q,K)$. Two independent MLPs read the \textsc{Line} token and produce raw revolute and prismatic direction vectors, $\hat{v}_{\mathrm{rev}} \in \mathbb{R}^3$ and $\hat{v}_{\mathrm{pris}} \in \mathbb{R}^3 $, respectively. Normalizing each vector gives $\hat{\mathbf a}_{\mathrm{rev}}$ and $\hat{\mathbf a}_{\mathrm{pris}}$. During inference, the predicted type $j$ selects the relevant head.
The fourth head predicts revolute axis offset relative to the query point conditioned on both the \textsc{Line} token and the predicted revolute axis direction $\hat{\mathbf a}_{\mathrm{rev}}$. To make this conditioning invariant to axis sign, we use outer-product embedding to embed the axis:
\begin{equation}
    \psi(\mathbf a)=
    [a_x^2,a_y^2,a_z^2,a_xa_y,a_xa_z,a_ya_z]^{\top}.
    \label{eq:axis_embedding}
\end{equation}
This embedding is continuous and satisfies $\psi(\mathbf a)=\psi(-\mathbf a)$. Separate MLPs project the \textsc{Line} token and $\psi(\hat{\mathbf a}_{\mathrm{rev}})$ to 64 dimensions. Their sum passes to a linear layer to produce a raw vector $\mathbf v_r\in\mathbb R^3$. We then enforce
\begin{equation}
    \hat{\mathbf r}=
    \left(\mathbf I_3-
    \hat{\mathbf a}_{\mathrm{rev}}\hat{\mathbf a}_{\mathrm{rev}}^{\top}\right)
    \mathbf v_r,
    \label{eq:offset_projection}
\end{equation}
where $\mathbf I_3$ is the identity matrix. This removes the component parallel to the axis and guarantees \cref{eq:hinge_offset}. Direction and location are coupled through both the conditioning embedding and this differentiable projection; the location loss, therefore, also updates the direction head.

\subsection{Losses}
\label{sec:approach:losses}

Joint type is supervised with a three-class cross-entropy
\begin{equation}
\mathcal L_{\mathrm{type}}=-\log p(j^*\mid I,\mathbf q,K)
\end{equation}
For each moving type $k\in\{\mathrm{rev},\mathrm{pris}\}$, we use an unsigned cosine loss to supervise the axis direction vector
\begin{equation}
    \mathcal L_{\mathrm{axis}}
    = 1-\left|\hat{\mathbf a}_{k}^{\top}\mathbf a^*\right|
    \label{eq:axis_regression_loss}
\end{equation}
For a revolute joint, let $\mathbf a^*$ be the unit ground-truth direction, $\mathbf o^*$ any metric point on its axis, and $z_q^*$ the ground-truth query depth. The normalized offset target is
\begin{equation}
    \mathbf r^*=
    \left(\mathbf I_3-\mathbf a^*\mathbf a^{*\top}\right)
    \left(\frac{\mathbf o^*}{z_q^*}-\tilde{\mathbf p}\right).
    \label{eq:normalized_offset_target}
\end{equation}
It is invariant to reversing $\mathbf a^*$, shifting $\mathbf o^*$ along its axis, and jointly scaling the metric scene and query depth. We supervise the normalized offset using Smooth L1 applied to the Euclidean prediction error:
\begin{equation}
\mathcal L_{\mathrm{offset}} =
\operatorname{SmoothL1}_{\beta}
\!\left(\|\hat{\mathbf r}-\mathbf r^*\|_2\right),
\qquad \beta=0.02.
\end{equation}

The base objective is
\begin{equation}
    \mathcal L=
    \mathcal L_{\mathrm{type}}
    +2\mathcal L_{\mathrm{axis}}^{\mathrm{rev}}
    +2\mathcal L_{\mathrm{axis}}^{\mathrm{pris}}
    +2\mathcal L_{\mathrm{offset}}.
    \label{eq:full_loss}
\end{equation}
Each term is averaged over its eligible examples, with empty terms set to zero. Eligibility depends on ground-truth labels, not predicted type, so an incorrect type prediction cannot suppress geometric supervision.

\subsection{Implementation}
\label{sec:approach:training}

The DINOv3 encoder is frozen; only the query encoder, decoder, and output heads are trained, for a total of 4.7M learnable parameters. We initialize these from scratch and train with AdamW on one NVIDIA RTX5090 GPU, with a learning rate of $10^{-4}$, weight decay of $5\times10^{-4}$, and a cosine schedule decaying to $10^{-6}$ over at most 15 full epochs. The optimizer batch size is $64$, dropout is $0.1$, and gradients are clipped to norm $5$. Learned computations use bfloat16 mixed precision, while geometric normalization, projection, and losses use float32. We train for at least four full epochs, then run early stopping with a patience of three. 

\section{Experiments}
\label{sec:experiments}

\begin{table*}[t]
\vspace{2mm}
\centering
\caption{Held-out split results on four datasets. Articulation-macro means grouped by GT
joint type. $R/P$: revolute/prismatic; axis error in degrees, recall in \%, $d_h$ and RMSE
in meters. $\dagger$: native trajectory protocol.}
\label{tab:indomain_baselines}
\begingroup
\fontsize{8.0}{9.6}\selectfont
\setlength{\tabcolsep}{1.3pt}
\renewcommand{\arraystretch}{1.12}
\setlength{\aboverulesep}{1.6pt}
\setlength{\belowrulesep}{1.6pt}
\setlength{\heavyrulewidth}{0.6pt}
\setlength{\lightrulewidth}{0.35pt}

\newcommand{\arttablepanel}[2]{%
  \begin{minipage}[t]{0.487\textwidth}
  \centering
  {\bfseries #1\par}\vspace{3pt}
  \begin{tabular*}{\linewidth}{@{}l@{\extracolsep{\fill}}r@{\extracolsep{0pt}\hspace{5pt}}r@{\extracolsep{\fill}}r@{\extracolsep{0pt}\hspace{5pt}}r@{\extracolsep{\fill}}r@{\extracolsep{\fill}}r@{\extracolsep{0pt}\hspace{5pt}}r@{\extracolsep{\fill}}r@{}}
  \toprule
  & \multicolumn{2}{c}{Axis error $\downarrow$}
  & \multicolumn{2}{c}{Type recall $\uparrow$}
  & $d_h\downarrow$ & \multicolumn{2}{c}{RMSE $\downarrow$} & $S\uparrow$ \\
  \cmidrule(lr){2-3}\cmidrule(lr){4-5}\cmidrule(lr){7-8}
  Method & \multicolumn{1}{c}{$R$} & \multicolumn{1}{c}{$P$} & \multicolumn{1}{c}{$R$} & \multicolumn{1}{c}{$P$} & \multicolumn{1}{c}{(m)} & \multicolumn{1}{c}{$R$} & \multicolumn{1}{c}{$P$} & \multicolumn{1}{c}{(\%)} \\
  \midrule
  #2
  \bottomrule
  \end{tabular*}
  \end{minipage}%
}

\arttablepanel{Articulate3D~\cite{Halacheva_2025_ICCV}}{%
OPDFormer-C~\cite{sun2024opdmulti} & 16.07 & 22.18 & 61.11 & 4.76 & 0.874 & 0.315 & 0.083 & 6.30 \\
MOPD~\cite{li2024mopd} & 11.94 & --- & \underline{87.37} & 0.00 & 0.823 & 0.281 & --- & 11.53 \\
A3VLM~\cite{pmlr-v270-huang25b} & 36.34 & 77.26 & 7.85 & 0.12 & 0.768 & 0.318 & 0.270 & 0.00 \\
3DOI~\cite{Qian_2023_ICCV} & \textbf{5.13} & \textbf{3.30} & 67.01 & \underline{86.05} & \underline{0.297} & \underline{0.091} & \textbf{0.012} & \underline{48.54} \\
\addlinespace[2pt]
VidBot~\cite{Chen_2025_CVPR} & --- & --- & --- & --- & --- & 0.302$^{\dagger}$ & 0.304$^{\dagger}$ & --- \\
MolmoMotion~\cite{zhang2026molmomotion} & --- & --- & --- & --- & --- & 0.106$^{\dagger}$ & 0.166$^{\dagger}$ & --- \\
\addlinespace[2pt]
\textbf{\ours (Ours)} & \underline{6.12} & \underline{9.38} & \textbf{96.20} & \textbf{89.25} & \textbf{0.201} & \textbf{0.062} & \underline{0.035} & \textbf{70.03} \\
}\hfill
\arttablepanel{MultiScan~\cite{mao2022multiscan}}{%
OPDFormer-C~\cite{sun2024opdmulti} & \textbf{2.47} & 28.38 & \underline{87.14} & 24.09 & \textbf{0.171} & \textbf{0.062} & 0.102 & \underline{42.66} \\
MOPD~\cite{li2024mopd} & \underline{3.28} & \underline{26.67} & 87.11 & 21.06 & \underline{0.196} & 0.072 & \underline{0.094} & 39.83 \\
A3VLM~\cite{pmlr-v270-huang25b} & 33.10 & 48.30 & 12.38 & 0.86 & 0.540 & 0.199 & 0.174 & 1.28 \\
3DOI~\cite{Qian_2023_ICCV} & 3.32 & 52.51 & 73.95 & \underline{37.41} & 0.326 & 0.104 & 0.183 & 23.93 \\
\addlinespace[2pt]
VidBot~\cite{Chen_2025_CVPR} & --- & --- & --- & --- & --- & 0.273$^{\dagger}$ & 0.213$^{\dagger}$ & --- \\
MolmoMotion~\cite{zhang2026molmomotion} & --- & --- & --- & --- & --- & 0.122$^{\dagger}$ & 0.204$^{\dagger}$ & --- \\
\addlinespace[2pt]
\textbf{\ours (Ours)} & 3.83 & \textbf{13.32} & \textbf{95.00} & \textbf{94.00} & 0.208 & \underline{0.068} & \textbf{0.049} & \textbf{60.12} \\
}

\par\vspace{10pt}

\arttablepanel{SceneFun3D~\cite{delitzas2024scenefun3d}}{%
OPDFormer-C~\cite{sun2024opdmulti} & 10.93 & \underline{15.78} & 82.48 & 36.68 & 0.554 & 0.157 & \underline{0.056} & 30.29 \\
MOPD~\cite{li2024mopd} & \underline{9.95} & 16.40 & \textbf{90.02} & 42.62 & 0.502 & 0.135 & 0.059 & 28.93 \\
A3VLM~\cite{pmlr-v270-huang25b} & 54.54 & --- & 5.96 & 0.00 & 0.740 & 0.246 & --- & 0.16 \\
3DOI~\cite{Qian_2023_ICCV} & 18.41 & 16.33 & 39.19 & \underline{69.17} & \underline{0.414} & \underline{0.127} & 0.058 & \underline{33.32} \\
\addlinespace[2pt]
VidBot~\cite{Chen_2025_CVPR} & --- & --- & --- & --- & --- & 0.152$^{\dagger}$ & 0.111$^{\dagger}$ & --- \\
MolmoMotion~\cite{zhang2026molmomotion} & --- & --- & --- & --- & --- & 0.135$^{\dagger}$ & 0.237$^{\dagger}$ & --- \\
\addlinespace[2pt]
\textbf{\ours (Ours)} & \textbf{7.51} & \textbf{6.56} & \underline{89.41} & \textbf{95.59} & \textbf{0.236} & \textbf{0.068} & \textbf{0.025} & \textbf{77.17} \\
}\hfill
\arttablepanel{PartNet-Mobility (synthetic)~\cite{xiang2020sapien}}{%
OPDFormer-C~\cite{sun2024opdmulti} & 40.47 & 29.20 & 49.27 & \underline{33.81} & 0.620 & 0.239 & 0.107 & 4.05 \\
MOPD~\cite{li2024mopd} & 36.49 & \underline{16.61} & \underline{57.48} & 22.08 & 0.761 & 0.294 & \underline{0.061} & \underline{7.28} \\
A3VLM~\cite{pmlr-v270-huang25b} & 54.64 & 61.84 & 46.10 & 23.55 & 0.524 & 0.216 & 0.216 & 3.21 \\
3DOI~\cite{Qian_2023_ICCV} & \underline{15.48} & 52.51 & 14.55 & 25.40 & \underline{0.471} & \underline{0.148} & 0.182 & 5.48 \\
\addlinespace[2pt]
VidBot~\cite{Chen_2025_CVPR} & --- & --- & --- & --- & --- & 0.179$^{\dagger}$ & 0.206$^{\dagger}$ & --- \\
MolmoMotion~\cite{zhang2026molmomotion} & --- & --- & --- & --- & --- & 0.185$^{\dagger}$ & 0.189$^{\dagger}$ & --- \\
\addlinespace[2pt]
\textbf{\ours (Ours)} & \textbf{5.18} & \textbf{3.61} & \textbf{96.52} & \textbf{93.81} & \textbf{0.078} & \textbf{0.031} & \textbf{0.013} & \textbf{89.67} \\
}

\endgroup
\end{table*}

\subsection{Data Generation}
\label{sec:experiments:data}

We train our model on five sources: three real-world datasets, Articulate3D~\cite{Halacheva_2025_ICCV}, SceneFun3D~\cite{delitzas2024scenefun3d}, and MultiScan~\cite{mao2022multiscan}, and two synthetic ones, PartNet-Mobility~\cite{xiang2020sapien} and HSSD~\cite{khanna2024hssd}. The real-world datasets provide articulation annotations and posed RGB frames, with depth obtained from sensors or reconstructed geometry; for the synthetic datasets, we render RGB-D frames and extract joint parameters from the simulator. We sample frames containing visible articulated parts and apply source-specific viewpoint-diversity checks. We convert source articulation annotations or simulator joint parameters into the camera coordinates of each sampled frame. Query points are sampled from visible part masks, with adaptive margins from mask boundaries, and retained only when their depth is valid and lies between 0.05 and 3\,m. For the real-world training datasets, we retain the source train, validation, and test splits.

Across all splits, Articulate3D contributes 2.6K views and 68K query points covering 1.3K distinct moving parts across 182 scenes.
SceneFun3D contributes 3.4K views and 74K query points covering 1K distinct moving parts across 176 scene visits.
MultiScan contributes 5K views and 146K query points covering 2.2K distinct moving parts across 227 scans of 96 environments. Our PartNet-Mobility renders contribute 30K views and 578K query points from 843 distinct object assets. Our HSSD renders contribute 22K views and 300K query points from 3K object instances, representing 1.5K distinct assets across 161 scenes. HSSD is used for training only; the held-out evaluations in \cref{tab:indomain_baselines} cover the remaining four sources.

To assess our model's generalization capabilities, we construct out-of-distribution evaluation sets from two real-world datasets, Arti4D~\cite{pmlr-v305-werby25a} and HOI!~\cite{engelbracht2025hoi}. We convert their annotations into the same representation and apply the same depth-validity checks, with additional manual annotation. These datasets provide additional real-world environments and viewpoints, and are used only for evaluation; they are excluded from our training data.

\subsection{Experimental Setup}
\label{sec:experiments:setup}

\textbf{Metrics and aggregation.}
For all experiments and all baselines, we average query errors across all views of each unique articulation, then average equally across articulations within each dataset. Axis error is $\arccos(|\hat{\mathbf a}^{\top}\mathbf a^*|)$ in degrees. For revolute joints, define the predicted line point nearest to the ground-truth query $\mathbf p$ as $\hat{\mathbf o}_q=\hat{\mathbf o}+\hat{\mathbf a}\hat{\mathbf a}^{\top}(\mathbf p-\hat{\mathbf o})$. The 3D hinge error is
\begin{equation}
e_h=\|(\mathbf I_3-\mathbf a^*\mathbf a^{*\top})(\hat{\mathbf o}_q-\mathbf o^*)\|_2.
\end{equation}
This error is invariant to the positions of the supplied origins along their respective lines.

As the end goal of this method is to generate trajectory motion for a robotic manipulator to execute, we also report the field RMSE, which measures the induced motion of up to 16 ground-truth surface query points on the same articulated part at $\{\pm10^\circ,\pm20^\circ,\pm30^\circ\}$ (revolute) or $\{0.1,0.2,0.3\}\,$m (prismatic). For $M$ points and $T$ states,
\begin{equation}
e_{\mathrm{RMSE}}=\min_{\sigma=\pm1}\!
\sqrt{\frac{1}{MT}\!\sum_{m,t}\!
\|\hat T_\sigma(t)\mathbf X_m-T^*(t)\mathbf X_m\|_2^2}.
\end{equation}
We report the hinge error and the field RMSE in meters, and type recall as a percentage. Like the axis error and type recall, the field RMSE is reported separately for ground-truth revolute and prismatic joints. All geometric errors are evaluated only on correctly classified examples with valid inputs for the corresponding metric.
We also report the overall success rate $S$, which requires a correct joint-type prediction, axis error $\leq10^\circ$, and field RMSE $\leq0.1\,\mathrm{m}$. Incorrect types and missing or invalid predictions count as failures. Success is computed for each query, averaged within each articulation, and then averaged equally across ground-truth moving articulations.

\textbf{Baselines.}
We compare against two types of baselines. The first set, OPDFormer-C~\cite{sun2024opdmulti}, MOPD~\cite{li2024mopd}, 3DOI~\cite{Qian_2023_ICCV}, and A3VLM~\cite{pmlr-v270-huang25b}, predicts articulation parameters and part segments from a single RGB image. The second set, MolmoMotion~\cite{zhang2026molmomotion} and VidBot~\cite{Chen_2025_CVPR}, predicts a point trajectory in 3D and requires additional privileged inputs. For methods that segment multiple parts within the image, we select the predicted articulation parameters whose mask overlaps the GT queries on the target object. We provide VidBot with GT 3D start and end points, and MolmoMotion with up to eight 3D query points and a language prompt. For both, we report only the RMSE metric.
For all baselines, we use their publicly released pretrained checkpoints. 

\subsection{Results}
\label{sec:experiments:results}

\begin{figure*}[t]
    \centering
    \includegraphics[width=\textwidth]{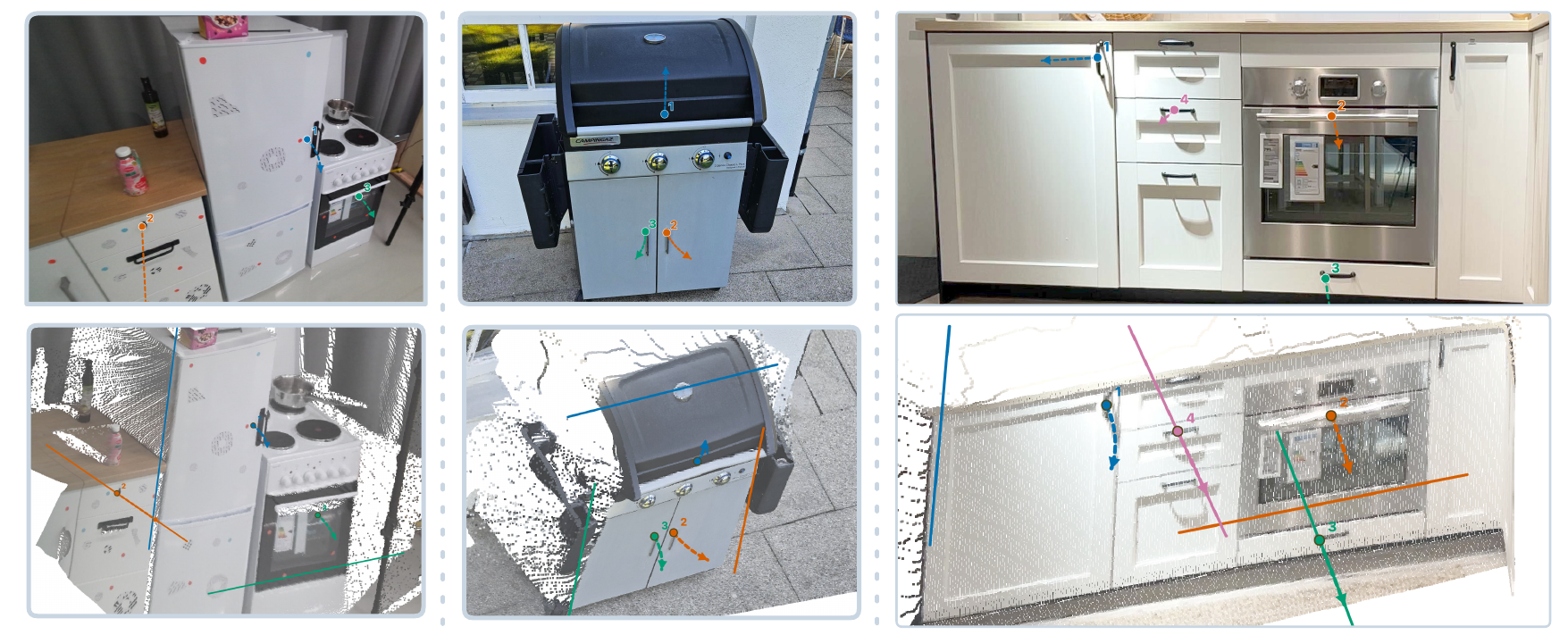}
    \caption{Qualitative articulation predictions on objects unseen during training:
Arti4D~\cite{pmlr-v305-werby25a} (left), self-captured scenes (middle), HOI!~\cite{engelbracht2025hoi} (right). Each scene is queried at several points. Top: input image with the query points and the projected motion trajectory. Bottom: the same predictions on the point cloud, with the predicted axis of motion and the 3D trajectory.}
    \label{fig:qualitative}
\end{figure*}

\textbf{In-distribution.}
\Cref{tab:indomain_baselines} reports results on four held-out dataset splits. \ours achieves the highest overall success rate on every dataset, ranging from $60.12\%$ on MultiScan to $89.67\%$ on PartNet-Mobility, and the highest prismatic type recall throughout. Among articulation-parameter methods, it also achieves the lowest hinge error on three datasets and the lowest field RMSE for both joint types on SceneFun3D and PartNet-Mobility. Some baselines remain stronger on individual conditional metrics: OPDFormer-C achieves lower revolute axis, hinge, and field RMSE errors on MultiScan, where MOPD and 3DOI also achieve a lower revolute axis error; 3DOI achieves lower axis errors for both joint types and lower prismatic field RMSE on Articulate3D; and MOPD achieves marginally higher revolute type recall on SceneFun3D. For VidBot~\cite{Chen_2025_CVPR} and MolmoMotion~\cite{zhang2026molmomotion}, neither of which predicts an explicit joint model, we report only the field RMSE under their native trajectory protocol ($\dagger$). \ours is more accurate than both on both joint types on all four datasets, despite both being trained on far larger amounts of general motion data. This suggests that a compact model trained for a specific motion type can be more accurate than a general trajectory prior while using far fewer trainable parameters. 

\begin{table*}[t]
\vspace{2mm}
\centering
\caption{Out-of-distribution results on HOI!~\cite{engelbracht2025hoi} and
Arti4D~\cite{pmlr-v305-werby25a}, both excluded from training. Columns and symbols as in
\cref{tab:indomain_baselines}.}
\label{tab:ood_baselines}
\begingroup
\fontsize{8.0}{9.6}\selectfont
\setlength{\tabcolsep}{1.3pt}
\renewcommand{\arraystretch}{1.12}
\setlength{\aboverulesep}{1.6pt}
\setlength{\belowrulesep}{1.6pt}
\setlength{\heavyrulewidth}{0.6pt}
\setlength{\lightrulewidth}{0.35pt}

\newcommand{\arttablepanel}[2]{%
  \begin{minipage}[t]{0.487\textwidth}
  \centering
  {\bfseries #1\par}\vspace{3pt}
  \begin{tabular*}{\linewidth}{@{}l@{\extracolsep{\fill}}r@{\extracolsep{0pt}\hspace{5pt}}r@{\extracolsep{\fill}}r@{\extracolsep{0pt}\hspace{5pt}}r@{\extracolsep{\fill}}r@{\extracolsep{\fill}}r@{\extracolsep{0pt}\hspace{5pt}}r@{\extracolsep{\fill}}r@{}}
  \toprule
  & \multicolumn{2}{c}{Axis error $\downarrow$}
  & \multicolumn{2}{c}{Type recall $\uparrow$}
  & $d_h\downarrow$ & \multicolumn{2}{c}{RMSE $\downarrow$} & $S\uparrow$ \\
  \cmidrule(lr){2-3}\cmidrule(lr){4-5}\cmidrule(lr){7-8}
  Method & \multicolumn{1}{c}{$R$} & \multicolumn{1}{c}{$P$} & \multicolumn{1}{c}{$R$} & \multicolumn{1}{c}{$P$} & \multicolumn{1}{c}{(m)} & \multicolumn{1}{c}{$R$} & \multicolumn{1}{c}{$P$} & \multicolumn{1}{c}{(\%)} \\
  \midrule
  #2
  \bottomrule
  \end{tabular*}
  \end{minipage}%
}

\arttablepanel{HOI!~\cite{engelbracht2025hoi}}{%
OPDFormer-C~\cite{sun2024opdmulti} & 21.69 & --- & \underline{93.79} & 0.00 & 0.471 & 0.170 & --- & 15.14 \\
MOPD~\cite{li2024mopd} & 15.90 & \underline{6.11} & 93.75 & 2.59 & 0.545 & 0.178 & \underline{0.023} & 18.70 \\
A3VLM~\cite{pmlr-v270-huang25b} & 51.76 & 88.13 & 4.44 & 0.41 & 0.570 & 0.231 & 0.300 & 0.00 \\
3DOI~\cite{Qian_2023_ICCV} & \textbf{9.64} & 6.90 & 69.50 & \underline{76.56} & \underline{0.241} & \underline{0.090} & 0.025 & \underline{58.58} \\
\addlinespace[2pt]
VidBot~\cite{Chen_2025_CVPR} & --- & --- & --- & --- & --- & 0.125$^{\dagger}$ & 0.074$^{\dagger}$ & --- \\
MolmoMotion~\cite{zhang2026molmomotion} & --- & --- & --- & --- & --- & 0.076$^{\dagger}$ & 0.179$^{\dagger}$ & --- \\
\addlinespace[2pt]
\textbf{\ours (Ours)} & \underline{10.55} & \textbf{5.89} & \textbf{98.21} & \textbf{77.84} & \textbf{0.170} & \textbf{0.059} & \textbf{0.022} & \textbf{69.08} \\
}\hfill
\arttablepanel{Arti4D~\cite{pmlr-v305-werby25a}}{%
OPDFormer-C~\cite{sun2024opdmulti} & 18.26 & 16.17 & 36.86 & 17.37 & 0.425 & 0.103 & 0.061 & 12.15 \\
MOPD~\cite{li2024mopd} & 37.73 & 15.98 & \underline{66.67} & 35.96 & 0.472 & 0.145 & 0.057 & 21.93 \\
A3VLM~\cite{pmlr-v270-huang25b} & 35.72 & 86.14 & 26.74 & 4.61 & 0.335 & 0.120 & 0.295 & 1.19 \\
3DOI~\cite{Qian_2023_ICCV} & \underline{10.94} & \underline{15.56} & 45.13 & \textbf{84.90} & \underline{0.232} & \underline{0.077} & \underline{0.055} & \underline{52.68} \\
\addlinespace[2pt]
VidBot~\cite{Chen_2025_CVPR} & --- & --- & --- & --- & --- & 0.107$^{\dagger}$ & 0.082$^{\dagger}$ & --- \\
MolmoMotion~\cite{zhang2026molmomotion} & --- & --- & --- & --- & --- & 0.097$^{\dagger}$ & 0.198$^{\dagger}$ & --- \\
\addlinespace[2pt]
\textbf{\ours (Ours)} & \textbf{6.55} & \textbf{3.82} & \textbf{91.73} & \underline{83.82} & \textbf{0.119} & \textbf{0.042} & \textbf{0.014} & \textbf{80.05} \\
}

\endgroup
\end{table*}

\textbf{Out-of-distribution.}
We evaluate generalization on HOI!~\cite{engelbracht2025hoi} and Arti4D~\cite{pmlr-v305-werby25a}, both excluded from training. As shown in \cref{tab:ood_baselines}, \ours achieves the highest overall success rates of $69.08\%$ and $80.05\%$, respectively. On correctly classified examples, its hinge errors are $0.170$ and $0.119\,\mathrm{m}$, with revolute/prismatic field RMSEs of $0.059/0.022\,\mathrm{m}$ and $0.042/0.014\,\mathrm{m}$. Among articulation-parameter methods, \ours achieves the best results on all reported metrics except revolute axis error on HOI!, where 3DOI performs better ($9.64^\circ$ versus $10.55^\circ$), and prismatic type recall on Arti4D, where 3DOI is slightly higher ($84.90\%$ versus $83.82\%$). These results demonstrate generalization to unseen object instances within the same broad categories under different viewing conditions, including the egocentric viewpoints of Arti4D. \Cref{fig:qualitative} illustrates predictions on unseen Arti4D, HOI!, and self-captured articulated objects.

\subsection{Real-Robot Experiments}
\label{sec:experiments:robot}

To test whether our model's predictions can drive real-world manipulation of articulated objects and generalize to unseen object instances, we design and conduct a real-world experiment on a mobile manipulation platform.

\textbf{Hardware Setup.}
We use a Boston Dynamics Spot equipped with an arm, and capture RGB images directly from its in-hand camera at a resolution of $640\times480$px. \ours is deployed on an NVIDIA Jetson Thor payload which runs on the back of the robot.%

\textbf{Experiment Setup.}
We run the experiment on common articulated furniture, such as drawers, cabinets, and doors, none of which the model saw during training. The experiments consist of 57 trials, spanning 16 articulated object parts, with viewpoints grouped into five classes: close, mid, far, left, and right.
For each object and view, we manually supply a candidate grasping point at which we query the model. This is typically the handle of the object, so that the same location serves both as the query pixel for the model and as the robot's grasp target. Combined with the metric depth from the in-hand depth camera at the query pixel, the prediction yields the metric-scale 3D joint parameters. The robot then grasps the handle using its own proprietary API, and executes the trajectory to interact with the part. The opening trajectory is directly extracted from \ours's prediction and followed using a Cartesian controller for the end-effector. A trial is counted as a success if the robot opens the part to the target range of motion: $20\,\text{cm}$ for prismatic joints and $45^\circ$ for revolute joints, and as a failure otherwise. All trials that were unsuccessful due to grasping failure were not scored and not counted in the final results. The remaining failure cases are directly attributable to wrong predictions.

\begin{figure}[tb]
    \centering
    \includegraphics[width=\columnwidth]{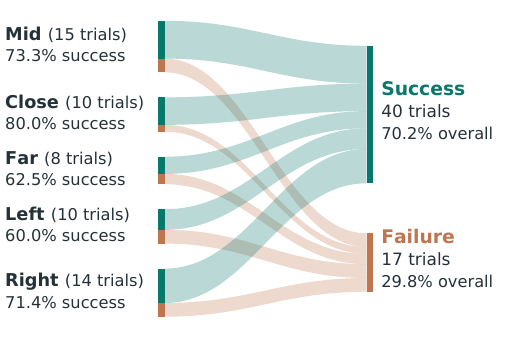}
    \caption{Real-robot trials by viewpoint. Each trial is assigned to one of five viewpoint classes and flows to a success or a failure outcome.}
    \label{fig:robot_viewpoints}
\end{figure}

\textbf{Results.}
Across the 57 trials, the robot succeeded in 40 and failed in 17, a 70.2\% success rate. \Cref{fig:robot_viewpoints} breaks the trials down by viewpoint class. We observe that the model's accuracy degrades when the viewpoint is at an extreme angle or when the camera cannot fully observe the object.

\subsection{Ablation Studies}
\label{sec:experiments:ablation}
\begin{table}[t]
\centering
\caption{Ablations on Articulate3D and HOI!~\cite{engelbracht2025hoi}. Full: absolute values;
other rows: signed change from Full, recall in percentage points. Units as in
\cref{tab:indomain_baselines}. \colorbox{green!15}{Green}: improvement;
\colorbox{red!15}{red}: regression.}
\label{tab:ablations}
\begingroup
\fontsize{7.4}{8.9}\selectfont
\setlength{\tabcolsep}{0.8pt}
\renewcommand{\arraystretch}{1.12}
\setlength{\aboverulesep}{1.2pt}
\setlength{\belowrulesep}{1.2pt}
\setlength{\heavyrulewidth}{0.6pt}
\setlength{\lightrulewidth}{0.35pt}
\begin{tabular*}{\columnwidth}{@{}l@{\extracolsep{\fill}}r@{\extracolsep{0pt}\hspace{5pt}}r@{\extracolsep{\fill}}r@{\extracolsep{0pt}\hspace{5pt}}r@{\extracolsep{\fill}}r@{\extracolsep{\fill}}r@{\extracolsep{0pt}\hspace{5pt}}r@{\extracolsep{\fill}}r@{}}
\toprule
& \multicolumn{2}{c}{Axis error $\downarrow$}
& \multicolumn{2}{c}{Type recall $\uparrow$}
& $d_h\downarrow$ & \multicolumn{2}{c}{RMSE $\downarrow$} & $S\uparrow$ \\
\cmidrule(lr){2-3}\cmidrule(lr){4-5}\cmidrule(lr){7-8}
Variant & \multicolumn{1}{c}{$R$} & \multicolumn{1}{c}{$P$} & \multicolumn{1}{c}{$R$} & \multicolumn{1}{c}{$P$} & \multicolumn{1}{c}{(m)} & \multicolumn{1}{c}{$R$} & \multicolumn{1}{c}{$P$} & \multicolumn{1}{c}{(\%)} \\
\midrule
\multicolumn{9}{@{}l}{\textbf{Articulate3D}} \\
\rowcolor{black!5}
\textbf{Full} & 6.12 & 9.38 & 96.20 & 89.25 & 0.201 & 0.062 & 0.035 & 70.03 \\
w/o pyramid & \cellcolor{red!15}+2.11 & \cellcolor{green!15}-1.49 & \cellcolor{green!15}+2.27 & \cellcolor{green!15}+0.24 & \cellcolor{red!15}+0.0270 & \cellcolor{red!15}+0.0074 & \cellcolor{green!15}-0.0056 & \cellcolor{red!15}-1.51 \\
dense attention & \cellcolor{green!15}-0.55 & \cellcolor{green!15}-0.07 & \cellcolor{green!15}+2.14 & \cellcolor{red!15}-1.95 & \cellcolor{red!15}+0.0603 & \cellcolor{red!15}+0.0137 & \cellcolor{green!15}-0.0003 & \cellcolor{red!15}-6.48 \\
w/o calibration & \cellcolor{red!15}+0.41 & \cellcolor{red!15}+0.46 & \cellcolor{red!15}-0.35 & \cellcolor{red!15}-1.71 & \cellcolor{red!15}+0.0177 & \cellcolor{red!15}+0.0045 & \cellcolor{red!15}+0.0017 & \cellcolor{red!15}-2.73 \\
metric + GT depth & \cellcolor{green!15}-0.17 & \cellcolor{red!15}+0.88 & \cellcolor{green!15}+2.43 & \cellcolor{red!15}-3.08 & \cellcolor{red!15}+0.0090 & \cellcolor{green!15}-0.0003 & \cellcolor{red!15}+0.0033 & \cellcolor{green!15}+1.24 \\
shared direction & \cellcolor{red!15}+0.75 & \cellcolor{red!15}+0.75 & \cellcolor{red!15}-0.19 & \cellcolor{green!15}+1.67 & \cellcolor{red!15}+0.0021 & \cellcolor{red!15}+0.0028 & \cellcolor{red!15}+0.0028 & \cellcolor{red!15}-0.13 \\
\midrule
\multicolumn{9}{@{}l}{\textbf{HOI! (OOD)}} \\
\rowcolor{black!5}
\textbf{Full} & 10.55 & 5.89 & 98.21 & 77.84 & 0.170 & 0.059 & 0.022 & 69.08 \\
w/o pyramid & \cellcolor{red!15}+0.39 & \cellcolor{red!15}+0.25 & \cellcolor{red!15}-0.53 & \cellcolor{red!15}-1.89 & \cellcolor{red!15}+0.0135 & \cellcolor{red!15}+0.0029 & \cellcolor{red!15}+0.0008 & \cellcolor{red!15}-1.34 \\
dense attention & \cellcolor{green!15}-1.80 & \cellcolor{red!15}+2.49 & \cellcolor{red!15}-1.14 & \cellcolor{green!15}+1.90 & \cellcolor{red!15}+0.0556 & \cellcolor{red!15}+0.0119 & \cellcolor{red!15}+0.0093 & \cellcolor{red!15}-5.80 \\
w/o calibration & \cellcolor{green!15}-1.95 & \cellcolor{red!15}+0.49 & \cellcolor{red!15}-1.55 & \cellcolor{green!15}+6.23 & \cellcolor{red!15}+0.0179 & \cellcolor{red!15}+0.0027 & \cellcolor{red!15}+0.0016 & \cellcolor{green!15}+2.45 \\
metric + GT depth & \cellcolor{red!15}+0.09 & \cellcolor{red!15}+0.13 & \cellcolor{red!15}-0.56 & \cellcolor{green!15}+0.43 & \cellcolor{red!15}+0.0165 & \cellcolor{red!15}+0.0024 & \cellcolor{red!15}+0.0005 & \cellcolor{red!15}-0.63 \\
shared direction & \cellcolor{red!15}+1.36 & \cellcolor{red!15}+1.99 & \cellcolor{red!15}-1.56 & \cellcolor{red!15}-3.20 & \cellcolor{red!15}+0.0199 & \cellcolor{red!15}+0.0058 & \cellcolor{red!15}+0.0073 & \cellcolor{red!15}-4.58 \\
\bottomrule
\end{tabular*}
\endgroup
\end{table}

To further understand the contribution of key components in our model, we conducted an ablation study on Articulate3D and HOI!~\cite{engelbracht2025hoi}. As shown in~\cref{tab:ablations}, removing the feature pyramid increases the revolute axis error and the hinge localization error on both datasets, showing that the model relies on the multiscale feature map to locate the hinge.
Replacing deformable cross-attention with dense cross-attention reduces the revolute axis error on both datasets but increases the hinge error and the revolute field RMSE, indicating that deformable attention contributes more to localizing the hinge than to orienting the axis. Removing the learned camera-context embedding improves revolute axis error and prismatic recall on the out-of-distribution dataset, yet increases the hinge error on both.

Predicting the metric offset $\hat{\mathbf r}$ directly, and recovering $\hat{\mathbf o}$ from it using the ground-truth query depth, is worse on most metrics. A single view fixes the scene only up to a global scale, so the metric offset is not identifiable from the image and has to come from a learned size prior. Normalizing by the query depth removes that dependence: at inference the measured depth multiplies both the query ray and the offset, so the measurement sets the scale exactly, whereas a directly regressed metric offset retains whatever scale was learned during training. Consistently with this, the degradation concentrates on the hinge error, while the scale-free axis direction is not consistently affected. Finally, replacing the two direction heads with a single head shared by both joint types also degrades most metrics, indicating that revolute and prismatic directions are better estimated separately. 

\section{Conclusion}
\label{sec:conclusion}

In this work, we presented \ours, a model for estimating articulation parameters from a single calibrated RGB image and a 2D query point. The model predicts the joint type, the 3D axis direction, and, for revolute joints, the 3D offset of the axis relative to the query point, expressed in units of the query point's depth. Unlike prior single-image methods, which first segment the moving part and regress metric 3D parameters directly, this formulation decouples \emph{where} to interact from \emph{how} the selected point moves; the depth-normalized representation removes the dependence on an absolute scale that a single view cannot resolve. This formulation aligns well with robotics settings where the robot must interact with unseen objects based on a single observation. 
Across four held-out splits of in-distribution data and two out-of-distribution datasets, \ours demonstrates robust performance and transfers to unseen object instances and viewpoints without retraining or fine-tuning. Real-robot trials further show that the predicted parameters can be scaled to metric units from a single depth reading, applied to previously unseen furniture, and used to successfully manipulate it. 

\section{Acknowledgements}

\footnotesize{This work has been supported by the German Federal Ministry of Research, Technology, and Space (BMFTR) under the Robotics Institute Germany (RIG).}

\begin{footnotesize}
    \bibliographystyle{IEEEtran}
    \bibliography{main.bib}
\end{footnotesize}

\end{document}